\pdfoutput=1
\documentclass[10pt,twocolumn,letterpaper]{article}

\usepackage{cvpr}
\usepackage{times}
\usepackage{epsfig}
\usepackage{graphicx}
\usepackage{amsmath}
\usepackage{amssymb}
\usepackage{float}

\usepackage[breaklinks=true,bookmarks=false]{hyperref}

\cvprfinalcopy 

\def\cvprPaperID{****} 

\ifcvprfinal\fi
\begin{document}

\title{Foreground-aware Image Inpainting}

\author{Wei Xiong$^{1}$\thanks{Work was primarily done while Wei Xiong was an Intern at Adobe.} \hspace{0.1in} Jiahui Yu$^2$ \hspace{0.1in} Zhe Lin$^3$ \hspace{0.1in} Jimei Yang$^3$ \hspace{0.1in} Xin Lu$^3$ \hspace{0.1in} Connelly Barnes$^3$ \hspace{0.1in} Jiebo Luo$^1$ \\
$^1$University of Rochester \hspace{0.2in}  $^2$University of Illinois at Urbana-Champaign \hspace{0.2in} $^3$Adobe Research \\
$^1${\tt\small \{wxiong5,jluo\}@cs.rochester.edu} $^2${\tt\small {jyu79@illinois.edu}}\\
$^3${\tt\small \{zlin, jimyang, xinl, cobarnes\}@adobe.com}
}

\maketitle
\thispagestyle{empty}

\begin{abstract}
	Existing image inpainting methods typically fill holes by borrowing information from surrounding pixels. They often produce unsatisfactory results when the holes overlap with or touch foreground objects due to lack of information about the actual extent of foreground and background regions within the holes. These scenarios, however, are very important in practice, especially for applications such as the removal of distracting objects. To address the problem, we propose a foreground-aware image inpainting system that explicitly disentangles structure inference and content completion. Specifically, our model learns to predict the foreground contour first, and then inpaints the missing region using the predicted contour as guidance. We show that by such disentanglement, the contour completion model predicts reasonable contours of objects, and further substantially improves the performance of image inpainting. Experiments show that our method significantly outperforms existing methods and achieves superior inpainting results on challenging cases with complex compositions.
  
\end{abstract}

\section{Introduction}

\begin{figure*}[htp]
  \setlength{\linewidth}{\textwidth}
  \centering
  \includegraphics[width=\textwidth]{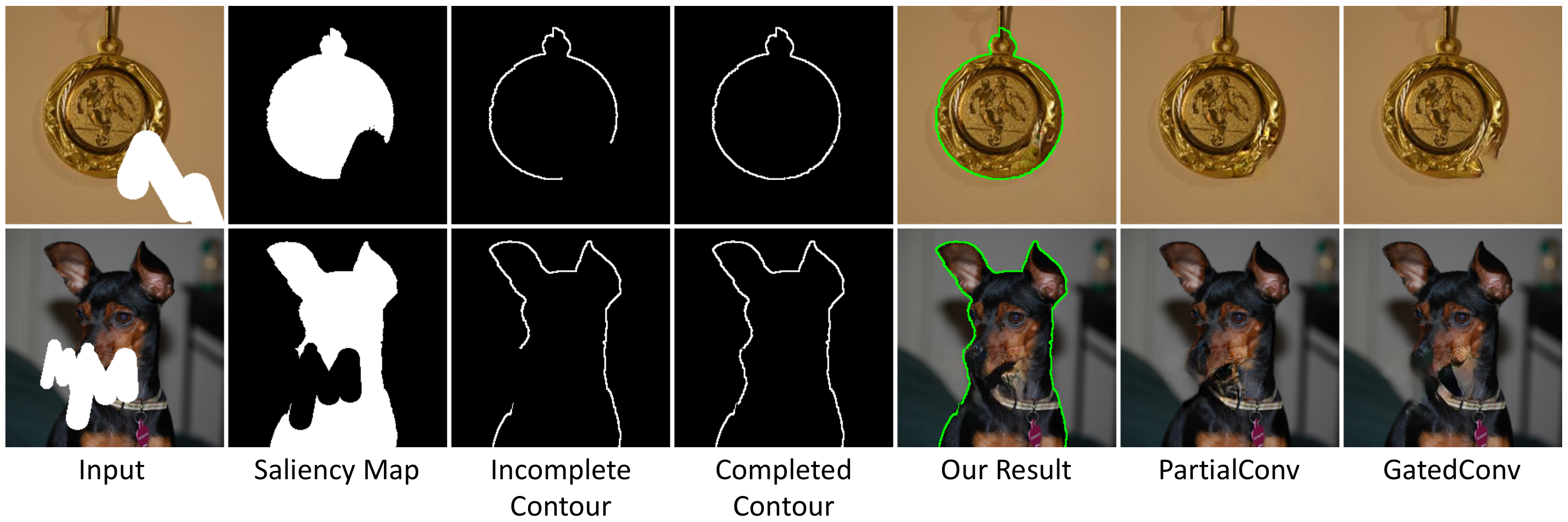}
  \caption{Our results compared with PartialConv \cite{liu2018image} and GatedConv \cite{yu2018free}. From left to the right are: image with holes, saliency map, incomplete contour and completed contour generated by our model, inpainting result of our model with completed contour (the green curve) on it, result of PartialConv \cite{liu2018image}, result of GatedConv \cite{yu2018free}, respectively.}
  \label{fig.comparison}
 \end{figure*}

Image inpainting is an important problem in computer vision, and has many applications including image editing, restoration and composition. We focus on hole filling tasks encountered commonly when removing unwanted regions or objects from photos. Filling holes in images with complicated foreground and background composition is one of the most significant and challenging scenarios.


Conventional inpainting methods~\cite{efros1999texture, bertalmio2003simultaneous, barnes2009patchmatch, simakov2008summarizing} typically fill missing pixels by matching and pasting patches based on low level features such as mean square difference of RGB values or SIFT descriptors~\cite{lowe2004distinctive}. These methods can synthesize plausible stationary textures but often produce critical failures in images with complex structures. To alleviate the problem, different structures of images have been exploited~\cite{huang2014image, hung2008exemplar, pnevmatikakis2008inpainting}. For example, Huang \etal~\cite{huang2014image} explicitly utilize planar structures as guidance to rectify perspectively-distorted textures. 
However, these methods still rely on existing patches and low-level features, and thus are unable to handle challenging cases where holes overlap with or are close to foreground objects. In such cases, a higher understanding of image content is required.

Recently, deep learning based methods~\cite{iizuka2017globally, liu2018image, yu2018free, yu2018generative, dekel2018sparse} have emerged as a promising alternative avenue by treating the problem as learning an end-to-end mapping from masked input to completed output. These learning-based methods are able to hallucinate novel contents by training on large scale datasets~\cite{karras2017progressive, zhou2017places}. To produce visually realistic results, generative adversarial networks (GANs)~\cite{goodfellow2014generative} are employed  to train the inpainting networks. However, by default all these methods assume that a generative network can learn to predict or understand the structure in the image implicitly, without explicit modeling of structures or foreground/background layers in the learning process. 

However, this has not been an easy task even for state-of-the-art models, such as PartialConv~\cite{liu2018image} and GatedConv~\cite{yu2018free}. For example, Fig.~\ref{fig.comparison} shows two common failure cases. On the top case, both GatedConv ~\cite{yu2018free} and PartialConv~\cite{liu2018image} fail to infer a reasonable contour in the missing region, and incorrectly predict a gold medal with an obvious notch. In addition, on the bottom case, both generate obvious artifacts around the neck of the dog. We conjecture that these failures may come from several limitations of current learning-based inpainting systems: (1) learning-based inpainting models are usually trained to fill randomly generated masks which are often completely located in the background or inside a foreground object. This is inconsistent with real-world cases where the holes might be close to or only have a small overlap with the foreground (\eg, cases of distracting region removal); (2) without explicitly modeling background and foreground layer boundaries, current deep neural network-based methods may not be able to predict the structure accurately inside the holes by simply training to fill random masks.

To this end, we propose a foreground-aware image inpainting system that explicitly incorporates the foreground object knowledge into the training process. Our system disentangles structure inference and image completion, and leverages accurate contour prediction to guide image completion. Specifically, our model first detects a foreground contour of the corrupted image, and then completes the missing contours of the foreground objects with a contour completion module. The completed contour along with the input image are then fed to the image completion module as guidance to predict contents in holes. 

The disentanglement of structure inference and image completion is conceptually simple and highly effective. Fig.~\ref{fig.comparison} shows that our model benefits greatly from the inferred contours. Our contour completion module is able to infer a reasonable structure in the missing region. Further, the image completion module takes predicted contours as guidance and generates cleaner contents around the borders of the objects. 

To summarize, our contributions are as follows: (1) We propose to explicitly disentangle structure inference and image completion to address challenging scenarios in image inpainting where holes overlap with or touch foreground objects. To the best of our knowledge, our work is among the first of a few studies that inpaint images with explicit contour guidance.  (2) To infer the structure of images, we propose a contour completion module trained explicitly to guide image completion. (3) To effectively integrate all the modules, we propose to adopt curriculum training on both the contour and image completion modules.  (4) Our experiments demonstrate that the system produces higher-quality inpainting results compared to existing methods.



\vspace{-0.1in}
\section{Related Work}
Image inpainting approaches can be roughly divided into two categories: traditional methods based on pixel propagation or patch matching, and recent methods based on deep neural network training. Traditional methods such as~\cite{ashikhmin2001synthesizing, ballester2001filling} fill in holes by propagating the neighborhood appearance based on techniques like isophote direction field. These methods are quite effective for small or narrow holes, but when the holes are large or the textures vary heavily, they often generate significant visual artifacts. Patch-based methods predict missing regions by searching for the most similar and relevant patches from the uncorrupted regions of the image. These methods work in an iterative way and can generate smooth and photo-realistic results, but at the cost of high computation cost and memory usage. To reduce the runtime and improve memory efficiency, tree-structure based search~\cite{mount1998ann} and randomized methods~\cite{barnes2009patchmatch} are proposed. PatchMatch~\cite{barnes2009patchmatch} is a typical patch based method that greatly speeds up the conventional algorithms and achieves high-quality inpainting results. A major drawback of PatchMatch lies in the fact that it searches for relevant patches from the whole image, without using any high-level information to guide the search. These methods work reasonably well for pure background inpainting tasks where holes are only surrounded by background textures, but could easily fail if holes overlap with an object or are close to an object. 

Recently, learning based inpainting methods \cite{liu2018image,yu2018generative,dekel2018sparse} have significantly improved inpainting results by learning semantics from large scale dataset. These methods typically train a convolutional neural network as a mapping function from a corrupted image to a completed one end-to-end. A significant advantage of these methods over the non-learning ones is the ability to learn and understand semantics of images for inpainting, which is especially important in cases of complex scenes, faces, objects and many others. Among these methods, Context Encoders is one of the first attempts~\cite{pathak2016context} that use a deep convolutional neural network to fill in the holes. It maps an image with a square hole to a complete image, and trains the model with L2 loss in the pixel space and an adversarial loss to generate sharper results. Similarly, Iizuka et al.~\cite{iizuka2017globally} use two discriminators to enforce that both the global appearance (whole image) and the local appearance ( content in hole) of the generated result are visually plausible. The method, however, still relies heavily on the post-processing of the completed image that blends both results from neural networks and traditional patch-matching methods. Yu et al.~\cite{yu2018generative} propose contextual attention to model long-range dependencies in images and a refinement network to eliminate post-processing, thus the whole system can be trained and tested end-to-end. However, these deep learning based inpainting methods typically infer the missing pixels conditioned on both valid pixels and the substitute values in the masked holes, which may lead to artifacts. Liu et al. ~\cite{liu2018image} address this problem by masking the convolution operation and updating the mask in each layer, so that the prediction of the missing pixels is only conditioned on the valid pixels in the original image. Yu et al.~\cite{yu2018free} further propose to learn the mask automatically with gated convolutions, and achieve better inpainting qualities. Additionally, Song et al.~\cite{songspg} apply a pretrained image segmentation network to obtain the foreground mask of the corrupted image, then fill the segmentation mask and use it to guide the completion of the image. However, these methods do not explicitly model the foreground and background boundaries. Therefore, they could fail in images where the masked region covers both foreground and background. 

\begin{figure*}[tbp]
	\centerline{\includegraphics[width=6.5in]{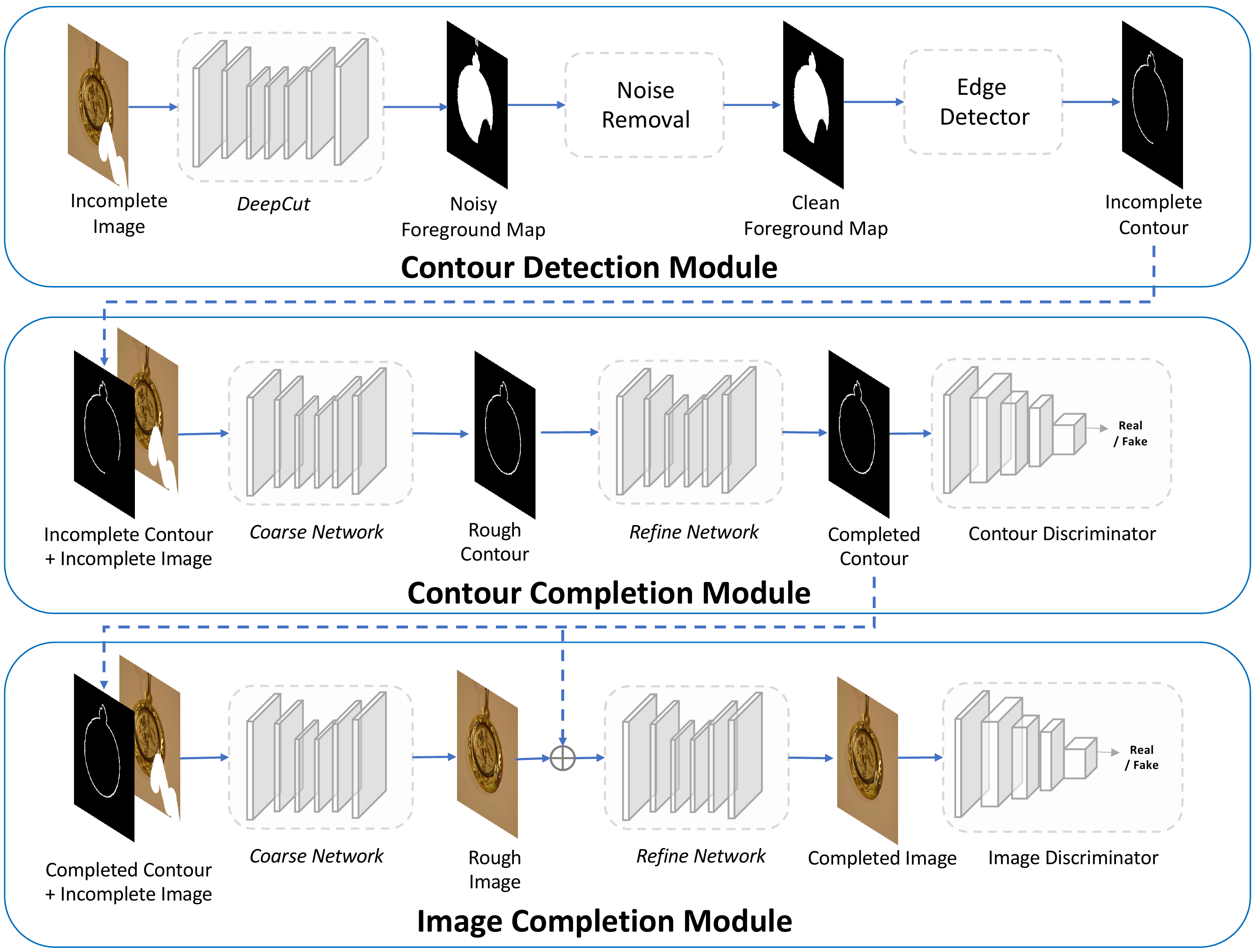}}
	\caption{The overall architecture of our inpainting model.}
	\label{fig.framework}
\end{figure*}

\section{Approach}

Given an incomplete image, our goal is to output a complete image with a visually pleasing appearance. The overall framework of our inpainting system is shown in Fig. \ref{fig.framework}. It is a cascade of three modules: incomplete contour detection module, contour completion module and image completion module. We automatically detect the contour of the incomplete image using the contour detection module. Then we use the contour completion module to predict the missing parts of the contour. Finally, we input both the incomplete image and the completed contour to the image completion module to predict the final inpainted image. To train our foreground-aware model, we need to prepare specific training samples and holes. In the following sections, we first introduce how we collect data and generate specific hole masks tailored to our task. Then we introduce the detailed implementation of our inpainting system.


\subsection{Data Acquisition and Hole Generation}

\textbf{Image Acquisition and Processing.} 
Existing datasets for image inpainting such as Places2 \cite{zhou2017places}, Paris \cite{Philbin08}, or CelebFace \cite{liu2015faceattributes} do not require any annotations, and training data pairs (image with hole and the ground-truth image) are typically constructed by generating random masks on the original images and by setting the original pixel values under the masks as the ground truth.
Our proposed framework for foreground-aware image inpainting requires us to train a contour completion module and infer the contour automatically, so we need a training dataset with labeled contours. One possibility is to directly use contour detection datasets, e.g. BSD500~\cite{amfm_pami2011}. However, such datasets are quite small in size and thus are not adequate to train an image inpainting model. Instead, we use salient object segmentation datasets as an alternative. We collect over 15,762 natural images that contain one or several salient objects, from a variety of public datasets, including MSRA-10K \cite{HouCvpr2017Dss}, manually annotated Flickr natural image dataset, and so on. Each image in this saliency dataset is annotated with an accurate segmentation mask. The dataset is quite diverse in content, containing a large variety of objects, including animals, plants, persons, faces, buildings, streets and so on. The relative size of objects in each image has a large variance, making the dataset quite challenging. We split all the samples into 12,609 training images and 3,153 testing images.

We then apply the Sobel edge operator on the segmentation mask to obtain the contours of the salient objects. Specifically, we first obtain the filtered mask $C_f$ by applying the Sobel operator: $C_f = |G_x| + |G_y|$, where $G_x$ and $G_y$ are the vertical and horizontal derivative approximations of the image, respectively. Then we binarize the filtered mask with a simple threshold and obtain the final binary contour  $C_{gt}$ as the ground-truth contour of the original image.

\textbf{Hole Mask Sampling.} In real-world inpainting applications, the distractors that users want to remove are usually arbitrarily-shaped, and usually not square-shaped. In order to simulate the real world inputs and learn a practical model, we draw holes on each image with arbitrary shapes randomly with a brush, based on the sampling method in \cite{yu2018free}. We generate two types of holes: 1). arbitrarily-shaped holes that can appear in any region of the input image. Under this setting, holes have a probability of overlapping with the foreground objects. This scenario is designed to handle the situations where unwanted objects are inside the foreground objects or partially occlude the salient objects; 2). arbitrarily-shaped holes that are restricted so that they have no overlap with the foreground objects. This type of holes are generated to simulate the situation where the unwanted regions or distracting objects are behind the salient objects. To deal with the second situation, we first randomly generate arbitrarily-shaped holes, then we remove the parts of holes that have overlap with the saliency objects. 

\subsection{Contour Detection}
During the inference stage, we do not have a contour mask of the input image. We therefore use DeepCut~\cite{deepcut} to detect the saliency objects in the image automatically. DeepCut uses a CNN-based architecture that extracts and combines high-level and low-level features to predict a salient object mask with accurate boundaries. Since the input image is corrupted with holes, the resulting segmentation map contains noise. In some situations, holes can even be treated as salient objects. To address this issue, we use the binary hole mask to remove the regions in the segmentation map that may be mistaken as salient objects. Then we apply connected component analysis \cite{samet1988efficient} to further remove some of the small clusters in the map to obtain the foreground mask. Then we adopt the Sobel Operator to detect the incomplete contour of the object from the segmentation map. The incomplete contour is then fed to the contour completion module to predict the missing contours. 

\subsection{Contour Completion Module}
The goal of our contour completion module is to complete the missing contours of the input image that are inside the hole regions. Given the incomplete image $I_{in}$, incomplete contour $C_{in}$ and the hole mask $H$ indicating the locations of the missing pixels, we aim to predict the complete contour  $C_c$ for the corrupted foreground objects. $C_c$ is a binary map with the same shape as the input image, with 1 indicating the boundary of the foreground objects and 0 for other pixels in the image. 
\vspace{-0.1in}
\subsubsection{Architecture}
The contour completion module is composed of a generator and a discriminator. The generator is a cascade of a coarse network and a refinement network. For training, instead of using predicted contours, we extract a clean incomplete contour $C_{in}$ of the foreground objects directly from the ground-truth contour $C_{gt}$ with the hole mask $H$, i.e., $C_{in} = H * C_{gt}$. Then we input the incomplete image, the incomplete contour image, and the hole mask into our coarse network, which outputs a coarse complete contour $C_{c}^{cos}$. The coarse network is an encoder-decoder network with several convolutional and dilated convolutional layers. The coarse contour map is a rough estimate of the missing contours. The predicted contours around the holes can be blurry and cannot be used as an effective guidance for the image completion module. 

To infer a more accurate contour, we adopt the refinement network which takes the coarse contour as input, and output a cleaner and more precise contour $C_{c}^{ref}$. The refinement network has a similar architecture as the coarse network, except that we use a contextual attention layer~\cite{yu2018generative}, to explicitly attend on global feature patches while inferring the missing values. Note that the pixel value of the predicted contour $C_{c}^{ref}$ ranges from 0 to 1, indicating the probability that the pixel to be on the actual contour. 

The refined contour is then fed to the contour discriminator for adversarial training. The contour discriminator is a fully convolutional PatchGAN discriminator \cite{isola2017image} that outputs a score map instead of a single score, so as to tell the realism of different local regions of the generated contour mask. Unlike discriminators for images, we discover that if we only input the contour mask (generated or ground-truth) to the discriminator, the adversarial loss is hard to optimize and the training tends to fail. This may be due to the sparse nature of the contour data. Unlike the natural images which have an understandable distribution on every local region, the pixels in the contour mask is sparsely distributed and contain less information for the discriminator to judge whether the generated distribution is close to the ground-truth distribution or not. 

To address this issue, we propose to adopt the ground-truth image as an additional condition, and use the image and contour pair as inputs to the contour discriminator. With this setup, the generated contour is not only required to be similar to the ground-truth contour, but also required to align with the contour of the image. The discriminator then obtains adequate knowledge to tell the difference between the generated distribution and the real distribution, and the training becomes stable. 
\vspace{-0.1in}
\subsubsection{Loss Functions}

To train the contour completion module, we will minimize the distance between the generated contour map $C_{c}^{cos}$, $C_{c}^{ref}$ and the ground-truth contour map $C_{gt}$. A straightforward way is to minimize the L1 or L2 distance between the masks in raw pixel space. However, this is not very effective as the contours in the mask are sparse, leading to the data imbalance problem. Determining the proper weights of each pixel is difficult. To address this issue, we propose to make use of the inherent nature of the contour mask, i.e., each pixel in the mask can be interpreted as the probability that the pixel is a boundary pixel in the original image. Therefore we can take the contour map as samples of a distribution, and calculate the distance with the ground-truth contour by calculating their binary cross-entropy between each pixel. We then adopt a focal loss~\cite{lin2018focal} to balance the importance of each pixel. Since our primary goal is to complete the missing contours, we pay more attention to the pixels in the holes by assigning them a larger weight. We formulate this loss as the content loss for contour completion $L_{con}^{C}$. The final loss function for the coarse contour is:

\begin{equation}
\small
	\begin{split}
	&\mathcal{L}_{con}^{C} \left( C_{c}^{cos}, C_{gt}\right) \\
	&= \frac{\lambda}{N}  \sum_{p}\left( H[p] ( C_{c}^{cos}[p]- C_{gt}[p])^2\mathcal{L}_{e} (C_{c}^{cos}[p], C_{gt}[p]) \right) \\
    &+ \frac{1}{N} \sum_{p}\left((1-H[p]) (C_{c}^{cos}[p]- C_{gt}[p])^2 \mathcal{L}_{e} (C_{c}^{cos}[p], C_{gt}[p]) \right)  \, .
    \end{split}
    \label{eq.1}
\end{equation}
where [p] denotes to the pixel spatial location of the contour map, $N$ is the number of pixels in the contour map, $\mathcal{L}_e(x, y)$ is the binary cross-entropy loss function, $x$ and $y$ are predicted probability score and the ground-truth probability, respectively.

Similarly, we use a content loss for the refined contour $ \mathcal{L}_{con}^{C} \left( C_{c}^{ref}, C_{gt}\right) $. The final content loss function for contour completion is:
\begin{equation}
	\mathcal{L}_{con}^{C} = \mathcal{L}_{con}^{C} (C_{c}^{cos}, C_{gt}) +  \mathcal{L}_{con}^{C} (C_{c}^{ref}, C_{gt})   \, .
\end{equation}

The focal loss helps to generate a clean contour. However, we observe that although we are able to reconstruct sharp edges in the uncorrupted regions, the contours in the corrupted regions are still blurry. To encourage the generator to produce sharp and clean contours, we use the contour discriminator $D^C$ to perform adversarial learning. Specifically, we use the recent technique called Spectral Normalization~\cite{miyato2018spectral} to stabilize the training of the GAN model. We use the hinge loss function to determine whether the input is real or fake. The adversarial loss for training the contour discriminator and the generator are as follows, respectively, where $\sigma$ denotes the ReLU function.
\begin{equation}
	\mathcal{L}_{adv}^{C} = \mathbb{E}[\sigma (1-D^C(C_{gt}))] + \mathbb{E}[\sigma (1 + D^C(C_{c}^{ref})] \, .
\end{equation}

\begin{equation}
	\mathcal{L}_{adv}^{C} = -\mathbb{E}[ D^C(C_{c}^{ref})] \, .
\end{equation}

\subsubsection{Curriculum Training}

Completing the contours is a challenging task. Although we have adopted a focal loss to balance the sparse data, and a spectral normalization GAN to obtain sharper results, we observe that it is still difficult to train the whole contour completion module. The training tends to fail if both the content loss and the adversarial loss are applied simultaneously even though the weights between the two types of losses are carefully adjusted. To avoid the issue, we use curriculum learning to gradually train the model. In the first stage, the contour completion module is required only to output a rough contour, thus we only train the model with the content loss. Then in the second stage, we fine-tune the pre-trained network with our adversarial loss, but with a very small weight compared to the content loss, i.e., 0.01 : 1 to avoid training failure due to the instability of the GAN loss for contour prediction. In the third stage, we fine-tune the whole contour completion module with the weight of adversarial loss and the weight of content loss to be 1:1. 

\subsection{Image Completion Module}

\subsubsection{Architecture}
Guided by the completed contours, our model gains the basic knowledge of where the foreground and background pixels are. This knowledge provides strong clues for the completion of the image. The image completion module takes the incomplete image $I_{in}$, the completed contour and the hole mask $H$ as inputs, and outputs the completed image $I_c$. It shares a similar architecture as the contour completion module. The generator of our image completion module also contains a coarse network and a refinement network. The coarse network outputs a coarsely completed image, which can be blurry with missing details. Then the refinement network takes the coarse image as input, and generates a more accurate result. 

By inputting both the incomplete image and completed contour to the coarse network, however, we observe that the final output of the generator tends to ignore the guidance of the completed contour. The shape of the generated image is not consistent with the input contour in the hole regions. This problem may be caused by the depth of the image completion networks. After layers of mapping, the knowledge provided by the completed contour can be forgotten or weakened, due to error accumulation. To tackle this problem, we input the completed contour to both the coarse network and the refinement network to enhance the effect of the condition. In this way, the effect of the contour condition can be stronger in the second stage of the image completion module. 

The discriminator takes the generated image/ground-truth image along with the hole mask indicating the location of the missing pixels as inputs, and tells whether the input pair is real or fake. Similar to the contour completion module, we use a PatchGAN structure and a hinge adversarial loss to train the model. 

\vspace{-0.1in}
\subsubsection{Loss Functions}
The loss function for the image completion module also consists of a content loss $\mathcal{L}_{con}^{I}$ and an adversarial loss $\mathcal{L}_{adv}^{I}$. The adversarial loss has a very similar form as the loss for contour completion, except that we apply the loss to the images instead of the contours. Note that the adversarial loss is only applied to the result of the refinement network. We do not apply the loss to the result of the coarse network. For the content loss, we use L1 loss to minimize the distance between the generated image and the ground-truth image. The image content loss is:

\begin{equation}
	\mathcal{L}_{con}^{I} =  \frac{1}{N}\sum_{p}\left( |I_c^{cos}[p] - I_{gt}[p]| +  |I_c^{ref}[p] - I_{gt}[p]| \right) \, .
	\label{eq.5}
\end{equation}
where $I_c^{cos}$, $I_c^{ref}$ and $I_{gt}$ are the output of the coarse network, the refinement network, and the ground-truth, respectively. [p] denotes the pixel spatial location of the image, $N$ is the number of pixels in the image.

\vspace{-0.1in}
\subsubsection{Training}

Our image completion module is first pre-trained on the large-scale Places2 dataset without the extra channel for the contour map, then fine-tuned on the saliency dataset with the guidance from the output of the contour completion module. Since the network we will fine-tune on the saliency dataset takes different inputs (takes additional contour as input) compared to the network we pretrain on the Places2 dataset, when fine-tuning our network, we keep the parameters of all the layers in the pretrained network except the first layer, and randomly initialize the first layers of our image completion module. To stabilize the training, we use a similar curriculum training strategy as the training of the contour completion module. 

There are two variations in our training process. The first one is to fix the parameters of the contour completion module, and only fine-tune the image completion module. The second way is to jointly fine-tune both modules. In our experiments, we observe that there are minor differences between these two so we fix our method as the second setting. 

\section{Experiments}

	
   

\begin{figure*}[tb]
\vspace{-0.2in}
	\centerline{\includegraphics[width=6.6in]{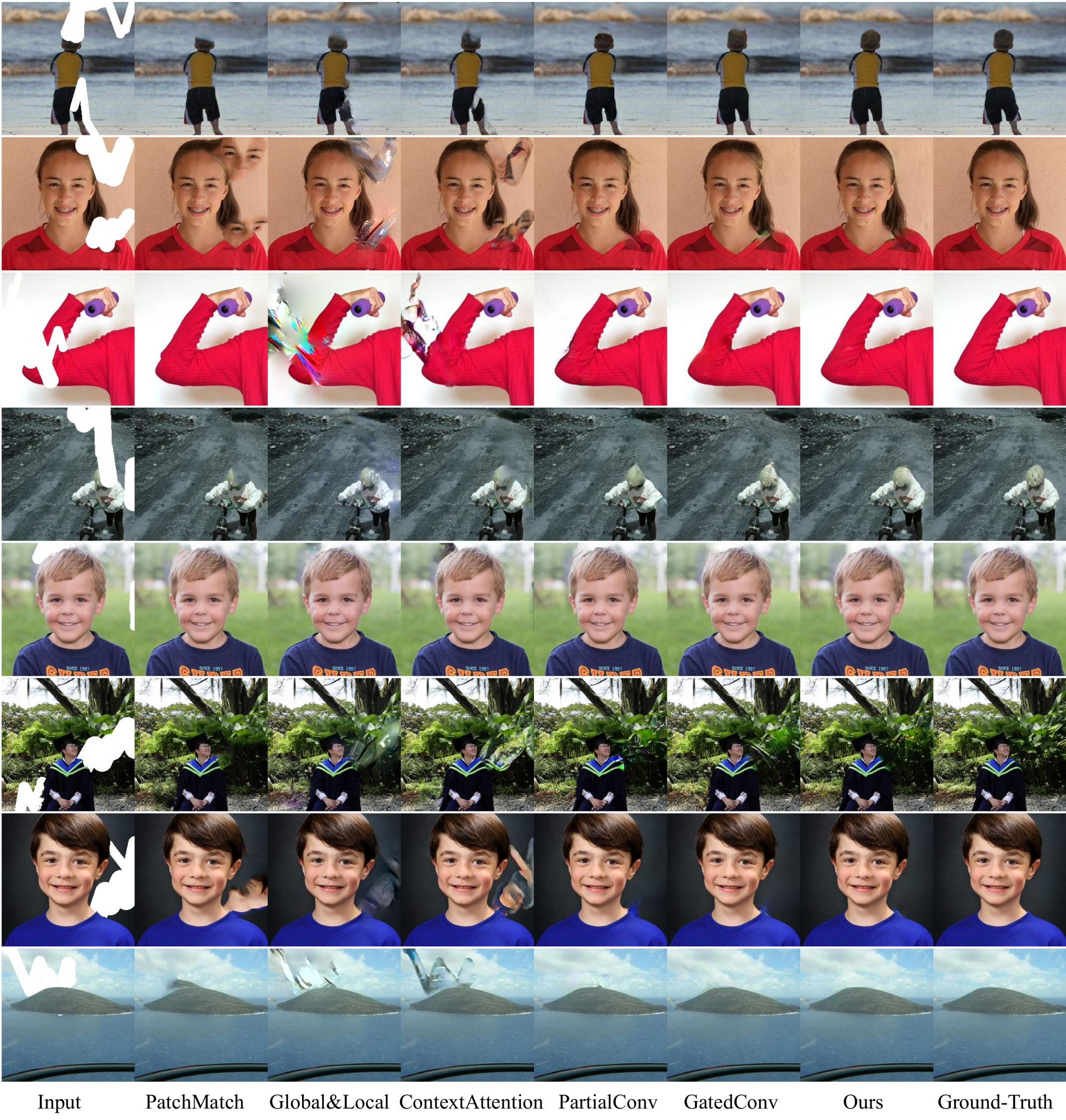} }
	\caption{Qualitative comparison between the state-of-the-art methods. Row 1-4 are samples with overlapped holes, while Row 5-8 are samples with non-overlapped holes. Please zoom in to see the details.}
	\label{fig.quality}
	\vspace{-0.2in}
\end{figure*}

\subsection{Implementation Details}
We obtain the incomplete contour of the foreground objects from our contour detection module and the pretrained DeepCut model \cite{deepcut}, without any finetuning. Then we train our contour completion module only on the saliency dataset. On the third stage, we first train the image completion module on the Places2, then finetune it on our saliency dataset. We also finetune both the contour completion module and the image completion module end-to-end on our saliency dataset. We use Adam as the optimizer, with a learning rate of 0.0002 and batchsize of 64 for both the contour completion module and the image completion module. $\lambda$ in Eq. \ref{eq.1} is set to 5 on training the contour completion module.


\subsection{Comparison with state-of-the-arts methods}
 In this part, we compare our proposed model with the state-of-the-art image inpainting methods on the validation set of our saliency dataset. We compare our full method (denoted as ``Ours Guided'') with  GatedConv~\cite{yu2018free}, PartialConv~\cite{liu2018image}, ContextAttention~\cite{yu2018generative}, Global\&Local~\cite{iizuka2017globally}, and PatchMatch~\cite{barnes2009patchmatch}. For a fair comparison, we also compare with GatedConv~\cite{yu2018free} fine-tuned on our saliency dataset, which can be regarded as the baseline - our model without contour prediction and guidance (denoted as ``Ours No Guide''). 

\vspace{-0.2in}
\subsubsection{Quantitative Evaluation}
 
We randomly select 500 images from the testing saliency dataset and generate both overlap and non-overlap holes for each image. Then we run each method on the corrupted images to obtain the final results. We use common evaluation metrics, i.e., L1, L2, PSNR, and SSIM, calculated using the complete image and the ground-truth image in pixel space, to quantify the performance of the models. Table \ref{table:quantitative} shows the evaluation results. Among the deep learning-based methods, our models outperform all the other methods in all four metrics. The results can be explained by that existing methods only consider making the textures of the completed image realistic, but ignore the structures of the image. Furthermore, our model with contour guidance brings consistent improvements over the baseline without guidance, demonstrating the validity of our proposed idea of leveraging contour prediction.

\vspace{-0.1in}
\subsubsection{Qualitative Evaluation}
Fig. \ref{fig.quality} shows visual comparisons of our method with existing methods. 
Seen from the figure, PatchMatch~\cite{barnes2009patchmatch} generates quite smooth textures. However, since it lacks an understanding of the image semantics, the generated image is not visually realistic when the holes are near the boundary of the foreground objects. Although Global\&Local~\cite{iizuka2017globally} and ContextAttention~\cite{yu2018generative} show the potential of handling holes with arbitrary shape (e.g., combining multiple small square holes to form an arbitrary shaped hole), since they are not specifically trained on arbitrary-shaped hole masks, they can generate artifacts which make the images unrealistic. PartialConv~\cite{liu2018image}, GatedConv~\cite{yu2018free} and our model without contour guidance (denoted as ``Ours No Guide'') can generate smooth and plausible images, but artifacts still exist around the borders of the objects. In addition, the shapes of the generated objects are not as natural as the real-world objects. Our full contour guided model not only generates a completed image with less artifacts, but also well completes the missing parts of the objects so that they have a very natural boundary. 


\vspace{-0.1in}
\subsubsection{User Study}

To make a more thorough evaluation of our method in terms of visual quality, we conduct a user study and show the result in Table \ref{table:user_study}. Specifically, we randomly select 50 images from our testing dataset, corrupt them with random holes and then obtain the inpainted results of each method. We show the results of each image to 22 users and ask them to select a single best result. Finally we collect 1,099 valid votes from all users. We count the number of times that each method is preferred by users. Table \ref{table:user_study} shows the user preferences of each method. Our full model is preferred the most, outperforming all the other methods by a large margin. This demonstrates the superiority of our foreground-aware model in terms of visual quality. 

\begin{table}[tbp]
 \vspace{-0.2in}
	\centering
    \small
	\renewcommand\arraystretch{1.1}
	
	\caption{Quantitative results on the saliency dataset.}
	\label{table:quantitative}
	\begin{tabular}{@{}lcccc@{}}
		\hline
		Method & L1 Loss & L2 Loss  & PSNR & SSIM \\
		\hline
		PatchMatch \cite{barnes2009patchmatch} &0.01386 &0.004278 &26.94 &0.9249  \\
        Global\&Local \cite{iizuka2017globally}&0.02450 &0.004445 &25.55 &0.9005  \\
        ContextAttention \cite{yu2018generative}&0.02116 &0.007417 &24.01 &0.9035 \\
        PartialConv \cite{liu2018image} &0.01085 &0.002437 &29.24 &0.9333 \\
        GatedConv \cite{yu2018free} &0.009966 &0.002531 &29.26 &0.9353 \\
        \hline
        Ours No Guided &0.010002 &0.002597 &29.35 &0.9356 \\
        Ours Guided&\textbf{0.009327} &\textbf{0.002329} &\textbf{29.86} & \textbf{0.9383}\\
		\hline
	\end{tabular}
   \vspace{-0.1in}
\end{table}

\begin{table}[tbp]
\vspace{-0.2in}
	\centering
	\small
	\renewcommand\arraystretch{1.1}
	
	\caption{User preference for the results of each method.}
	\label{table:user_study}
	\begin{tabular}{lc}
		\hline
        Method &  Preference Counts \\
        PatchMatch \cite{barnes2009patchmatch} &  23\\
        Global\&Local \cite{iizuka2017globally}& 5\\
        ContextAttention \cite{yu2018generative}& 4\\
        PartialConv \cite{liu2018image} & 90\\
        GatedConv \cite{yu2018free}& 100\\
        \hline
		Ours No Guide & 146\\
		Ours Guided & \textbf{731}\\
        \hline
        
	\end{tabular}
\end{table}

\subsection{Ablation Study}

\begin{figure}[tbp]
    \vspace{-0.1in}
	\centerline{\includegraphics[width=3.2in]{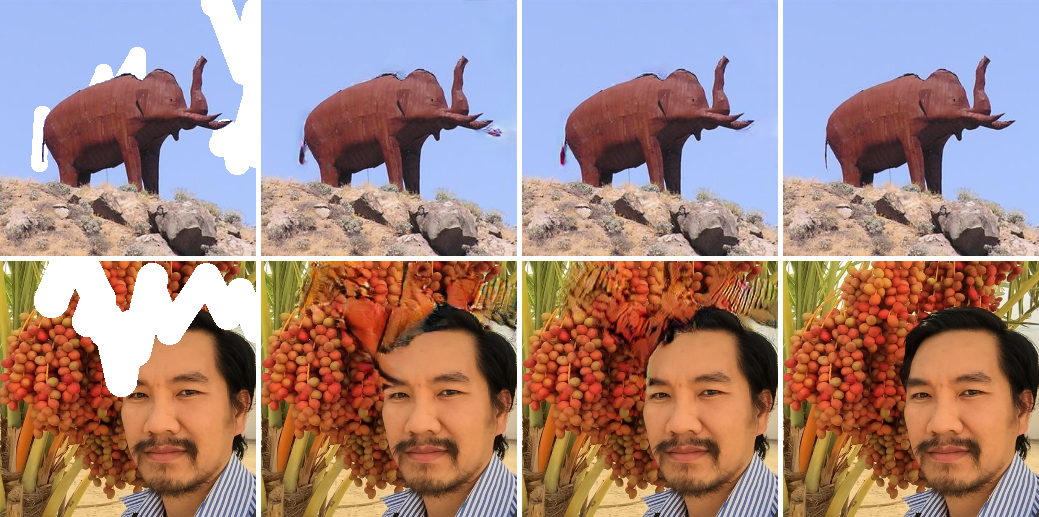}}
	\caption{Comparison between our model with/without contour as guidance. From left to right: input image with holes, our model without contour guidance, our full model, the ground-truth.}
	\label{fig.ablation}
    \vspace{-0.15in}
\end{figure}

We also analyze how our contour completion module contributes to the final performance of image inpainting. We compare our full model to the model without contour as guidance, as is shown in Fig. \ref{fig.ablation}. The top row shows the results where holes have no overlap with the foreground object, while the bottom shows the case where holes overlap with the object. In both cases, our model without contour guidance generates obvious artifacts around the border of the foreground object, while our model with contour guidance can infer object boundaries correctly and produce realistic inpainting results. The comparison indicates that the completed contours greatly improve the performance of the image inpainting model and that contour guidance is a crucial part to the success of our model. 
 
\vspace{-0.1in}
\section{Conclusion}
 In this paper, we propose the foreground-aware image inpainting model for challenging scenarios involving prediction of both foreground and background pixels. Our model first detects and completes the contours of the foreground objects in the image, then uses the completed contours as a guidance to inpaint the image. It is trained on a specifically collected saliency image dataset. Experiments show that our model can generate natural contours of objects, which are of great benefit for image completion. Our model significantly outperforms various state-of-the-art models both quantitatively and qualitatively. This shows that using structures to indicate the foregrounds and backgrounds of the input image, then explicitly guide the completion of the image is a promising direction for inpainting tasks. 
\vspace{-0.1in}
\section{Acknowledgements}
This work is partially supported by NSF awards \#1704309, \#1722847 and \#1813709.

{\small
\bibliographystyle{ieee_fullname}
\bibliography{main}
}

\end{document}


\title{Supplementary Material}

\maketitle
\thispagestyle{empty}

\section{Network Architecture}
In this section, we introduce the detailed configuration of our networks. Our model is composed of three modules, the contour detection module, the contour completion module and the image completion module. 

\subsection{Contour Detection Module}
The key component of our contour detection module is the saliency object segmentation network. We describe the details of this network here. The segmentation network used in our paper consists of three major parts: High-level Stream, Low-level Stream and Boundary Refine Module.

\textbf{High-level Stream} It takes the incomplete image as input and uses the encoder part of a traditional segmentation network to extract compact features. The output is a two-channel low resolution feature map, which is used as the bottle-neck of the network. In this module, we use Inception V2 as the segmentation network. The input of the network is a 3-channel image and the original output of the truncated Inceptions-V2 is a 7x7 1024-channel feature map. In order to get a 14x14 feature map, we use dilated convolution for the last two inception modules. Finally, we add a convolution layer to generate the 2-channel 14x14 feature map. 

\textbf{Low-level Stream} This module is a shallow network composed of a single 7x7 convolution layer with a stride of 1. The input to the shallow network is our incomplete image. The output of this stream is a 64-channel feature map that has the same spatial size as the input image. 

\textbf{Boundary Refine Module}  This module takes the low-level and high-level feature as input and outputs the final result. Specifically, we first resize the high-level feature map to the original resolution by bilinear upsampling. Then, we concatenate the upsampled high-level feature map with the low-level feature map and pass them to the densely connected layer units. Each dense unit is composed of some convolutional layers, and the output will be concatenated with the input to the unit.

\subsection{Contour Completion Module}
Our contour completion module shares a similar architecture with GatedConv \cite{yu2018free}. Specifically, it consists of two stages. The first stage is a encoder-decoder network that takes the incomplete contour, the incomplete image and the mask as inputs, and outputs a coarse result of the completed contour. The encoder is a cascade of several gated convolution blocks described in \cite{yu2018free}, and finally maps the input image to feature maps with a spatial resolution of 64x64. The decoder has a reverse architecture as the encoder and maps the feature maps to a completed contour image. The coarse contour is then concatenated with the mask and then input to the refine network of the contour completion module, to get the final result. The refine network has a two-stream encoder that maps the inputs to feature maps of size 64x64, and a decoder that maps the feature maps to the final image. 

The detailed configuration of the contour completion module is as follows. For simplicity, we denote kernel size, dilation, stride size and channel number as K, D, S, C, respectively.  

\textbf{Coarse Network:} K5S1C48 - K3S2C96 - K3S1C96 - K3S2C192 -
K3S1C192 - K3S1C192 - K3D2S1C192 - K3D4S1C192 -
K3D8S1C192 - K3D16S1C192 - K3S1C192 - K3S1C192 -
resize (2×) - K3S1C96 - K3S1C96 - resize (2×) - K3S1C48
- K3S1C24 - K3S1C3 - sigmoid.

\textbf{Refine Network:} 

Branch-1: K5S1C48 - K3S2C96 - K3S1C96 - K3S2C192 -
K3S1C192 - K3S1C192 - K3D2S1C192 - K3D4S1C192

Branch-2: K5S1C48 - K3S2C948 - K3S1C96 - K3S2C192 - K3S1C192 - K3S1C192 (contextual attention) - K3S1C192 - K3S1C192

Decoder: concat - K3S1C192 - K3S1C192 - resize (2×) - K3S1C96 - K3S1C96 - resize (2×) - K3S1C48 - K3S1C24 - K3S1C3 - sigmoid.

\begin{figure*}[tp]
	\centerline{\includegraphics[width=6.75in]{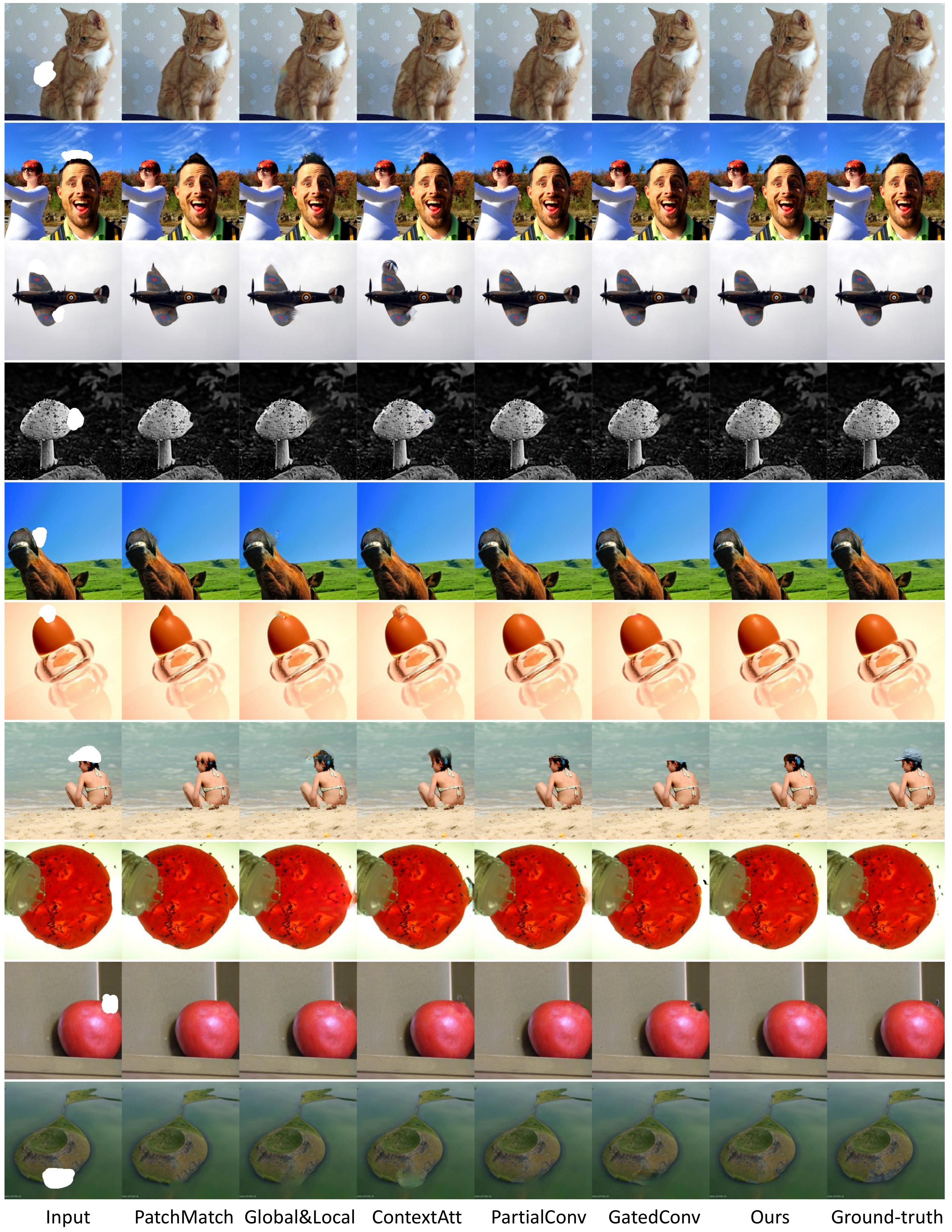} }
	\caption{Qualitative comparison between the state-of-the-art methods. Please zoom in to see the details.}
	\label{fig.quality}
\end{figure*}

\begin{figure*}[tp]
	\centerline{\includegraphics[width=6.75in]{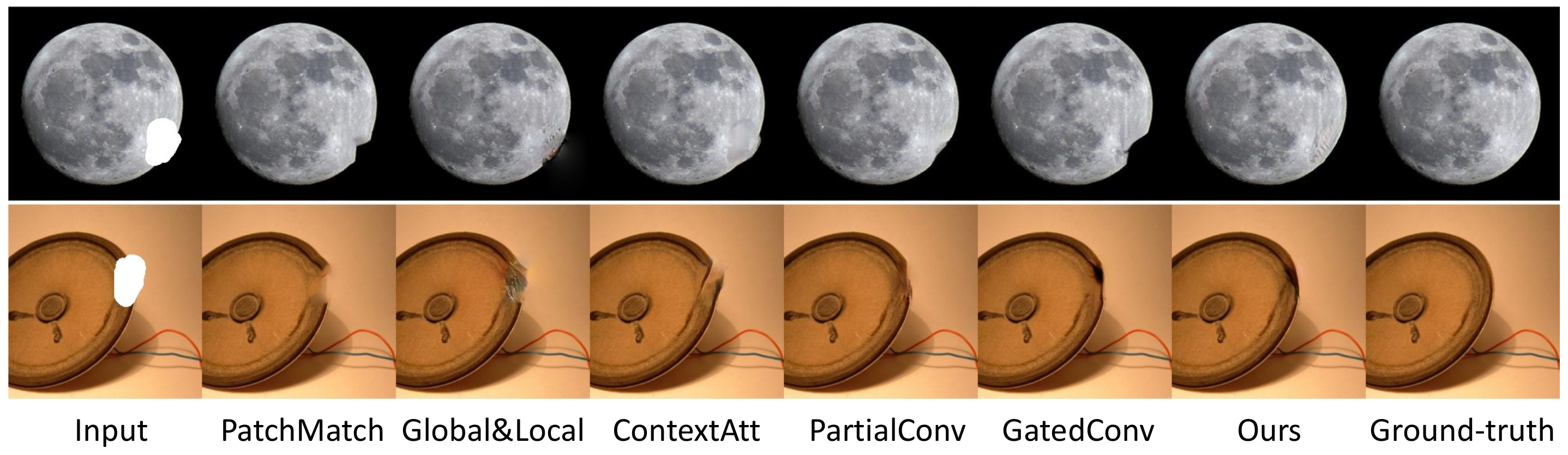} }
	\caption{Failure cases. Please zoom in to see the details.}
	\label{fig.failure}
\end{figure*}

\subsection{Image Completion Module}
The image completion module has the same architecture as the contour completion module, except for the inputs and outputs to each network. The input to the refine network is the coarse completed image, the completed contour and the mask, the output of the coarse network and the refine network is activated with tanh function, instead of sigmoid  which is used in the contour completion module.

\section{Comparison with State-of-the-arts}
\subsection{Qualitative Results}
In this section, we show more qualitative results. As can be seen from Fig. \ref{fig.quality}, our model consistently outperforms the state-of-the-art models.
\subsection{Quantitative Results}
To make a more thorough comparison, we also include the results for each model using Perceptual Similarity LPIPS \cite{zhang2018unreasonable} on the feature space of VGG or AlexNet, and the results are shown in Table \ref{table:1}. 

\begin{table}[h]
	\centering
	\renewcommand\arraystretch{1.1}
	
	\caption{Additional quantitative metrics, smaller is better.}
	\label{table:1}
	\begin{tabular}{l|c|c}
		\hline
		{Method } & LPIPS (VGG) & LPIPS (Alex)\\
		\hline
		PConv  & 0.064 & 0.044 \\ 
		GatedConv & 0.063  & 0.048\\
		Ours Guided & \textbf{0.060} & \textbf{0.043} \\
		\hline
	\end{tabular}
\end{table}

\section{Contour Completion Results}
\begin{figure}[tb]
	\centerline{\includegraphics[width=3.2in]{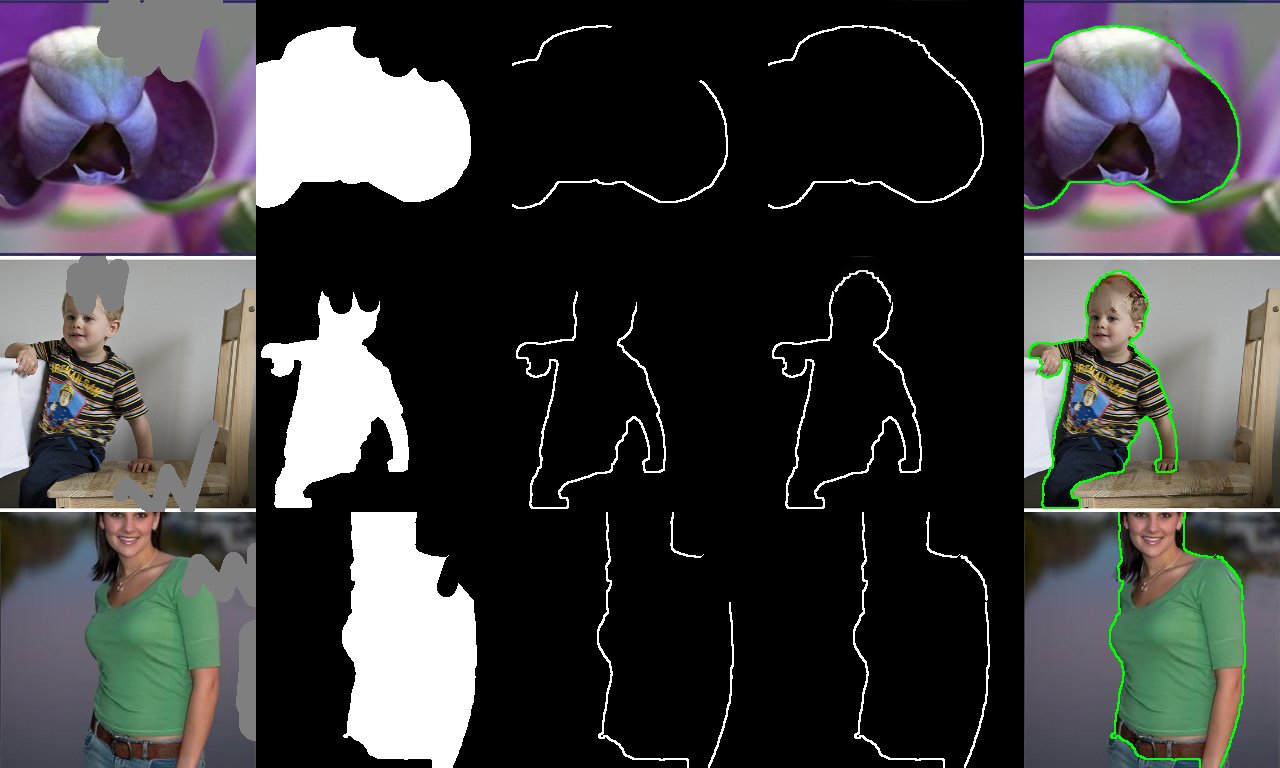}}
	\caption{Contour completion results. From left to right: image with hole, saliency map of the incomplete image, incomplete contour, completed contour and the completed image.}
	\label{fig.contour}
\end{figure}

We supplement more contour completion results here. As is shown in Fig. \ref{fig.contour}, our contour completion module can infer clean, sharp and reasonable contours, which can be of great benefits to the completion of the image. 

\section{Failure Cases}

In this section, we show some cases that the existing models fail to inpaint. The results are shown in Fig. \ref{fig.failure}. Seen from the figure, though our model is able to complete a reasonable contour for the incomplete object, however, sometimes, artifacts can still occur. In our future work, we will try to reduce the artifacts while predicting a reasonable shape for the objects.

{\small
\bibliographystyle{ieee_fullname}
\bibliography{supp}
}